\documentclass{article}
\usepackage{times}
\usepackage{amssymb,amsmath,bbm}
\usepackage{techrep}
\usepackage{epsfig}

\begin{document}

\title{Evolino for Recurrent Support Vector Machines}

\author{
J{\"u}rgen Schmidhuber\footnotemark[2] \footnotemark[3]\hspace*{5mm} Matteo Gagliolo\footnotemark[2]\hspace*{5mm}Daan Wierstra\footnotemark[2]\hspace*{5mm} Faustino Gomez\footnotemark[2] \\
\footnotemark[2] IDSIA Galleria 2, 6928 Manno (Lugano), Switzerland  \\
\footnotemark[3] TU Munich, Boltzmannstr. 3,  85748 Garching, M\"{u}nchen, Germany \\
{\tt \{juergen,matteo,daan,tino\}@idsia.ch}}
\reportnumber{19-05 (version 2.0)}
\date{11 October 2005, revised 15 December 2005}

\renewcommand{\thefootnote}{\fnsymbol{footnote}}
\makecover
\maketitle \renewcommand{\thefootnote}{\arabic{footnote}}

\begin{abstract}
Traditional Support Vector Machines (SVMs) need pre-wired  finite time
windows to predict and classify time series.  They do not have an
internal state necessary to deal with sequences involving arbitrary
long-term dependencies.  Here we introduce a new class of recurrent, truly
sequential SVM-like devices with internal adaptive states, trained by
a novel method called EVOlution of systems with KErnel-based outputs
(Evoke), an instance of the recent Evolino class of methods
\cite{Schmidhuber:05ijcai,Wierstra:05geccoevolino}.  Evoke evolves
recurrent neural networks to detect and represent temporal
dependencies while using quadratic programming/support vector
regression to produce precise outputs, in contrast to our recent work
\cite{Schmidhuber:05ijcai,Wierstra:05geccoevolino} which instead uses
pseudoinverse regression.  Evoke is the first SVM-based mechanism 
learning to classify a context-sensitive language. 
It also outperforms recent state-of-the-art gradient-based 
recurrent neural networks (RNNs) on various time series prediction tasks.
\end{abstract}

\section{\label{sc:intro}Introduction}

Support Vector Machines (SVMs) \cite{Vapnik:95} are powerful
regressors and classifiers that make predictions based on a linear
combination of kernel basis functions. The kernel maps the input
feature space to a higher dimensional space where the data is linearly
separable (in classification), or can be approximated well with a hyperplane
(in regression).  A limited way of applying existing SVMs to 
sequence prediction~\cite{mukherjee97nonlinear,uller97predicting} or
classification \cite{Salomon:02icslp} is to build a training set either by transforming the
sequential input into some static domain (e.g., a frequency and phase representation, 
a Hidden Markov model (HMM) \cite{Jaakkola:98nips,Jebara:04jmlr}, a simple frequency count of
symbols or substrings \cite{Lodhi:00nips}), 
or by considering restricted, fixed time windows of $m$ sequential input values. 
One alternative presented
in \cite{Shimodaira:02nips} is to average kernel distance between elements 
of input sequences aligned to $m$ points. 
Such window-based approaches are obviously bound to fail if there are temporal
dependencies exceeding $m$ steps; 
while HMMs present numerous local minima when trained with long sequences \cite{Bengio:95b,Hochreiter:01icann}.
In a more sophisticated approach by Suykens and Vandewalle \cite{Suykens:00},
a window of $m$ previous output values is fed back
as input to a recurrent model with a fixed kernel.
So far, however, there has not been any recurrent SVM that  {\em learns} to 
create internal state representations for sequence learning tasks involving
time lags of arbitrary length between important input events.
For example, consider the task of correctly classifying arbitrary instances of the context-free language $a^nb^n$
($n$ a's followed by $n$ b's, for arbitrary integers $n>0$).  
 
Our novel algorithm, EVOlution of systems with KErnel-based outputs
(Evoke),  addresses such problems. It evolves a 
recurrent neural network (RNN) as a preprocessor for a standard SVM kernel. 
The combination of both can be  viewed as an adaptive
kernel learning a task-specific distance 
measure between pairs of input sequences.
Although Evoke uses SVM methods, it
can solve several tasks that traditional SVMs cannot  even solve
in principle.  

Evoke is a special instance of a recent,
broader algorithmic framework for supervised sequence learning called
Evolino: EVolution of recurrent systems with Optimal LINear Output
\cite{Schmidhuber:05ijcai,Wierstra:05geccoevolino}.  Evolino combines
neuroevolution (i.e.\  the artificial  evolution of neural networks)
and analytical linear methods that are optimal according to various criteria.  The
underlying idea of Evolino is that often a linear model can account
for a large number of properties of a sequence learning problem.
Non-linear properties unpredictable by the linear model are then dealt
with by more general evolutionary optimization processes.  Recent work
has focused on the traditional problem of minimizing mean squared
error (MSE) summed over all time steps of a time series to be
predicted. An optimal linear mapping from hidden nodes to output nodes
was obtained through the  Moore-Penrose pseudoinverse 
method (i.e.\ PI-Evolino), which
is both fast and optimal in the sense that it minimizes MSE
~\cite{penrose:pseudo}.  The weights of the more complex, nonlinear
hidden units were found through evolution, where the the fitness
function was the residual error on a validation set, given the
training-set-optimal  linear mapping from hidden to output nodes.

In the present work we use a different optimality criterion, namely,
the maximum margin criterion of SVMs~\cite{Vapnik:95}.
Hence the optimal linear output weights are evaluated using
quadratic programming,
as in traditional SVMs, the difference here being  the evolutionary
RNN preprocessing of the input.

The resulting Evoke system not only learns to solve  tasks  unsolvable
by any traditional SVM, but also outperforms recent state-of-the-art
RNNs on certain tasks, including Echo State Networks (ESNs)
\cite{Jaeger:04} and previous gradient descent RNNs
\cite{Werbos:74,RumelMc:86,Williams:89,RobinsonFallside:87tr,Pearlmutter:95,Hochreiter:97lstm}.

\section{\label{sc:evoke}The Evoke Algorithm}

\begin{figure}[tb]
\begin{minipage}{0.5\linewidth}
\centerline{\epsfig{file=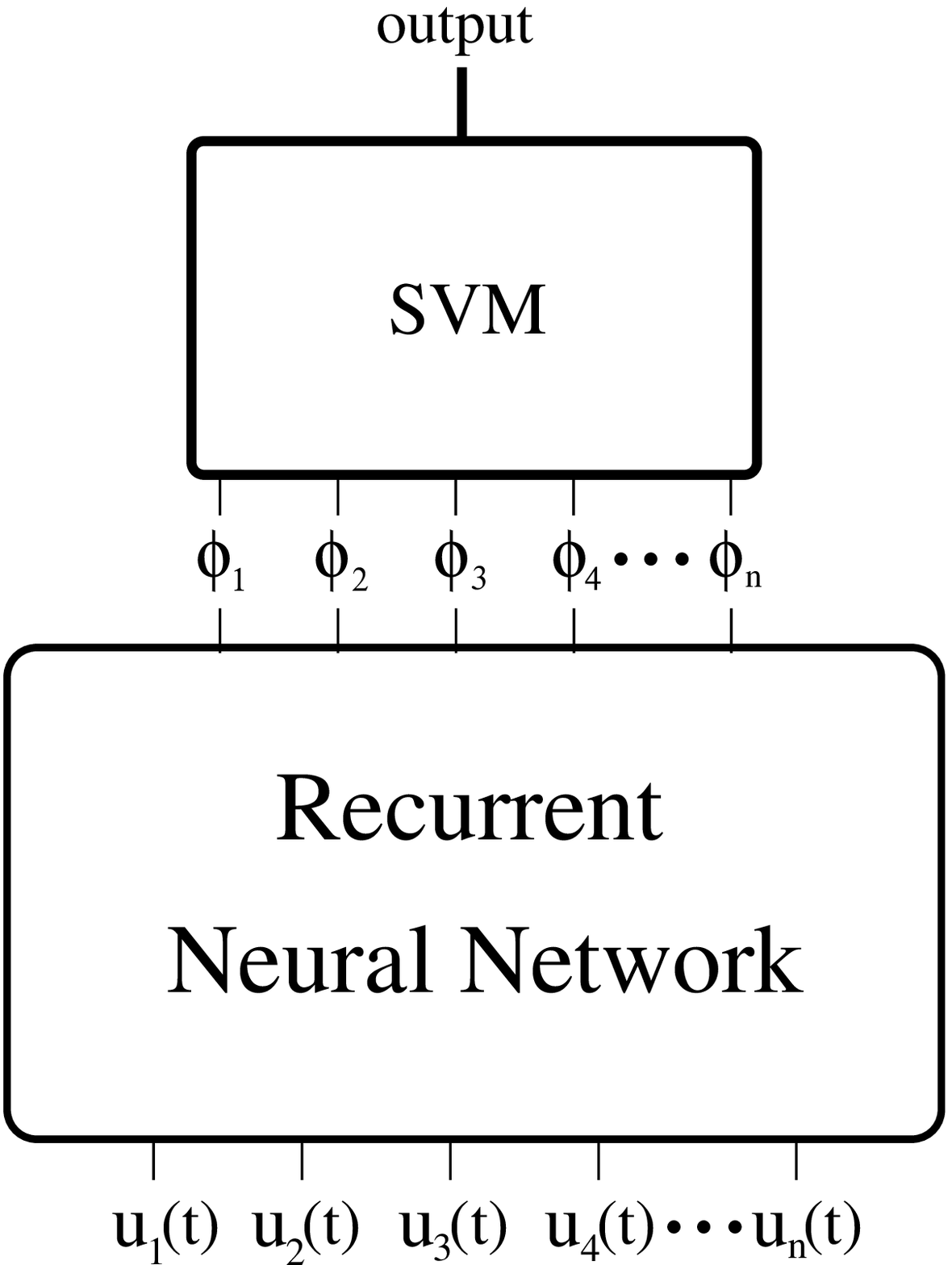, width=0.8\linewidth }}
\centerline{\footnotesize (a) }
\end{minipage}
\begin{minipage}{0.5\linewidth}
\centerline{\epsfig{file=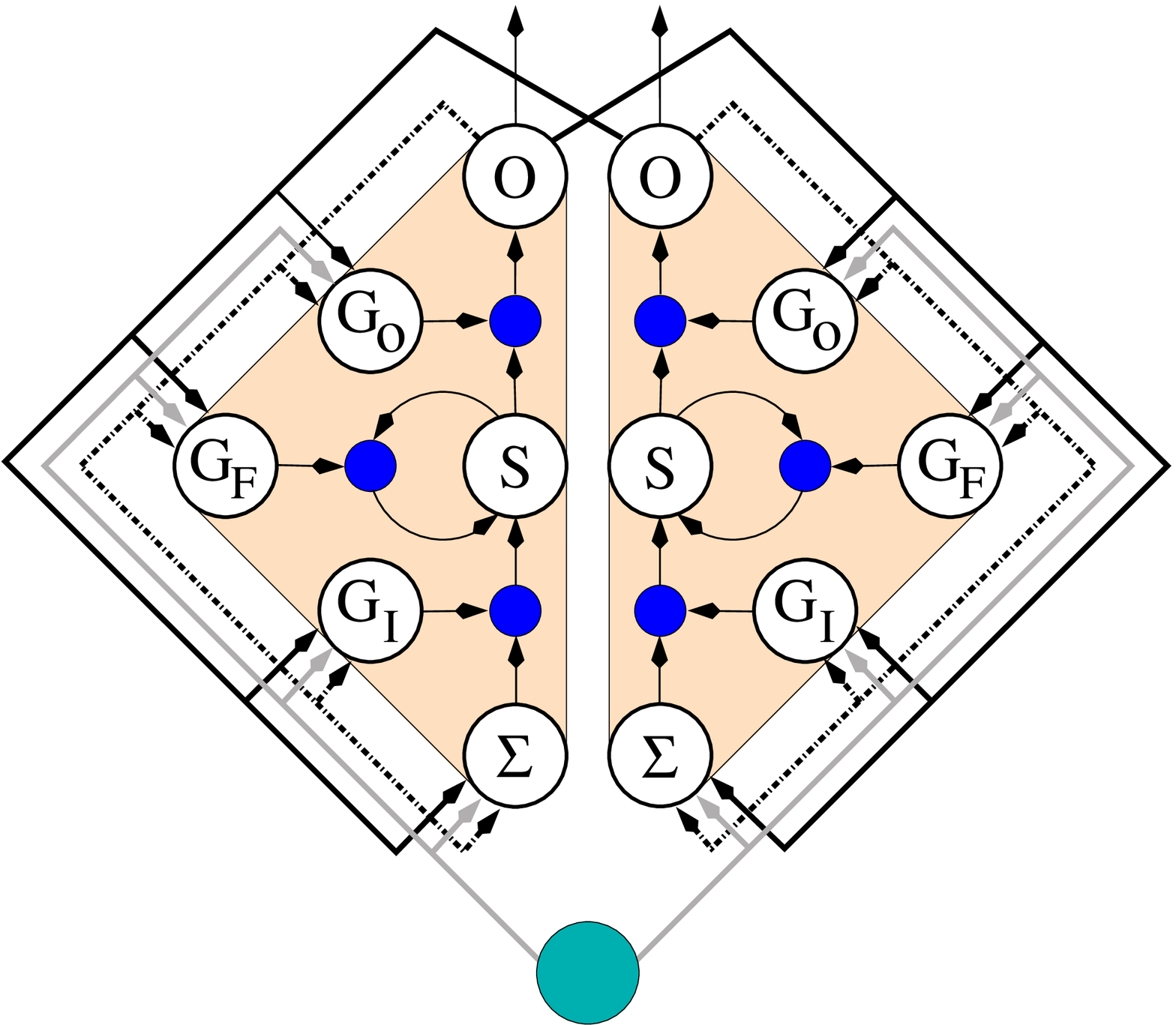, width=\linewidth }}
\centerline{\footnotesize (b) }
\end{minipage}
\caption{ (a) {\bf Evoke network.}  An RNN receives sequential inputs
$\mathbf{u}(t)$ and produces neural activation vectors $\phi_1 \ldots
\phi_n$ at every time step $t$. These values are  fed as input to a
Support Vector Machine, which outputs a scalar
$y(t)$. While the RNN is evolved, the  weights of the SVM module are
computed with 
support vector  regression/classification.
(b) {\bf Long Short-Term Memory.}  The figure shows the 
LSTM architecture that we use for the RNN module.  This example network 
has one input (lower-most circle),  and two {\em memory cells} (two
triangular regions).  Each cell has an internal state $S$ together
with a  Forget gate ($G_F$)  that determines how much the state is
attenuated  at each time step.  The Input gate ($G_I$) controls access
to the cell by the external inputs that are summed into each $\Sigma$
unit, and the  Output gate ($G_O$) controls when and how much the cell's
output unit ($O$) fires.  Small dark nodes represent the multiplication 
function.}
\label{fig:evoke-general}
\end{figure}

Evolino systems are based on two cascaded modules: 
(1) a recurrent neural network
that  receives the sequence of external inputs,  and (2) a parametric
function that  maps the internal activations of the first module to  a
set of outputs.  
In particular, an Evoke network (Figure~\ref{fig:evoke-general}a)
is governed by the following formulas:

\begin{equation}\label{eq:phit}
\boldsymbol{\phi}(t) =
f(\mathbf{W},\mathbf{u}(t),\mathbf{u}(t-1),\ldots,\mathbf{u}(0)),
\end{equation}
\begin{equation}\label{eq:yt}
y(t) = w_0 +\sum_{i=1}^k
\sum_{j=0}^{l_i}w_{ij}K(\boldsymbol{\phi}(t),\boldsymbol{\phi}^{i}(j)),
\end{equation}

\noindent where $\boldsymbol{\phi}(t)\!\in\!\mathbb{R}^n$ is the
activation at time $t$ of the $n$ units of the RNN, $f(\cdot)$, given the sequence of input vectors
$\mathbf{u}(0)..\mathbf{u}(t)$, and weight matrix
$\mathbf W$. Note that, because  the networks are recurrent, $f(\cdot)$ is
a function of the entire input history. The output $y(t)\!\in\!\mathbb{R}$ 
of the combined system can be interpreted as a class label, in classification
tasks, or as a prediction of the next input $\mathbf{u}(t+1)$,
in time-series prediction.
To compute $y(t)$ we take the weighted sum of
the kernel distance  $K(\cdot,\cdot)$
between $\boldsymbol{\phi}(t)$ and each activation vector $\boldsymbol{\phi}^i(j)$
obtained by first running the  training set of sequences through the network
(see below).  

In order to find a $\mathbf W$ that minimizes the error between $y(t)$ and the
correct output, we use 
artificial evolution~\cite{Rechenberg:71,Schwefel:74,Holland:75}.  
Starting with random population of real-numbered strings or 
{\em chromosomes} representing
candidate weight matrices,
we evaluate each candidate through the 
following two-phase procedure.

In the first phase, the aforementioned training set of sequence pairs, 
$\{\mathbf{u}^i, d^i\},
i=1..k$, each of length $l^i$,
is presented to the network.  For each input sequence $\mathbf{u}^i$,
starting at time $t=0$, each pattern $\mathbf{u}^i(t)$ is
successively propagated through the RNN to produce a vector of 
activations 
$\boldsymbol{\phi}^i(t)$
that is stored as a row in a $n\times \sum_i^k l^i$ 
matrix $\Phi$.   
Associated with each 
input sequence is a {\em target} row
vector $d^i$ in $D$ containing the correct output values for each time
step.   Once all $k$ sequences have been  seen,  the weights
$w_{ij}$  of the kernel model (equation \ref{eq:yt})  
are computed using support vector
regression/classification from $\Phi$ to $D$,
with $\{\boldsymbol{\phi}^i,d^i\}$ as training set.
 
In the second phase,
a validation set  is presented to  the network, but now the inputs are
propagated through the  RNN  {\em and} the newly computed
output connections to produce $y(t)$.  The error in the
classification/prediction  or  the {\em residual error}, possibly
combined with the error on the training set, is then used as the
fitness measure to be minimized by evolution.  By measuring error
on the validation set rather that just the training set, 
RNNs will receive better fitness for being able to generalize.

Those RNNs that are most fit are then selected
for reproduction where new candidate RNNs are created by exchanging 
elements between chromosomes and an possibly mutating them.
New individuals replace the worst old ones and the cycle repeats
until a sufficiently good solution is found.

This idea of evolving neural networks using artificial evolution or
{\em neuroevolution}~\cite{yao:review} 
is normally applied to reinforcement learning tasks
where correct network outputs (i.e.\ targets) are not known {\it a
priori}.  However, Evolino/Evoke uses it for supervised learning with
feedback based on a validation set (as opposed to the traditional
training set).  Instead of trying to evolve an RNN that makes predictions
directly, we use an RNN to perform a non-linear transformation
from the arbitrary-dimensional space of sequences to 
the finite-dimensional space of neural activations, where the 
SVM can operate.
This way we can exploit the powerful generalization capability of
SVMs, in the context of sequential data.

In this study, Evoke is instantiated   using Enforced SubPopulations
(ESP;~\cite{gomez:phd})  to evolve Long Short-Term Memory 
(LSTM;~\cite{Hochreiter:97lstm}) networks.  We combine these two particular
methods because both have routinely outperformed previous methods in
their domains
\cite{Gomez:03+,Gomez:05geccocern,Hochreiter:97lstm,Gers:2000nc,
Gers:01ieeetnn,Gers:02jmlr,Schmidhuber:02nc,Perez:03,graves:05nn}.

ESP  differs from standard  neuroevolution  methods in that, instead
of evolving complete networks, it {\em coevolves} separate
subpopulations of network components or {\em neurons}.  If the
performance of ESP does not improve for a predetermined number of
generations, a technique called {\em burst
mutation}\cite{gomez:phd,Schmidhuber:05ijcai} is used, to inject
diversity into the subpopulations.

LSTM  is an RNN purposely designed to learn  long-term dependencies
via gradient descent.  The unique feature of the LSTM architecture is
the {\em memory cell} that is capable of maintaining its activation
indefinitely 
(figure~\ref{fig:evoke-general}b).   Memory cells consist of a linear
unit  which holds the {\em state} of the cell, and three gates that
can open or close over time.  The Input gate ``protects'' a neuron
from its input: only when the gate is open, can inputs affect the
internal state of the neuron. The Output gate lets the internal state
out to other parts of the network,  and the Forget gate enables the
state to ``leak'' activity when it is no longer useful. The gates also
receive inputs from neurons, and a function over their input (usually
the sigmoid function) decides whether they open or close.
\cite{Hochreiter:97lstm,Gers:2000nc,Gers:01ieeetnn,Gers:02jmlr,
Schmidhuber:02nc,Perez:03,graves:05nn}.  
Hereafter, the term gradient-based LSTM (G-LSTM) 
will be used to refer to LSTM when it is trained in the 
conventional way using gradient-descent.

ESP and LSTM are combined by coevolving subpopulations of memory cells
instead of standard recurrent neurons.   Each chromosome is a string
containing the external input weights  and the Input, Output, and
Forget gate weights, for a total
of $4*(I+H)$ weights in each memory cell chromosome, where $I$ is the
number of external inputs and $H$ is the number of memory cells in the
network.  There are four sets of $I+H$  weights because the three
gates and the cell itself receive input from outside the cell
and the other cells.  ESP normally uses crossover to recombine
neurons.  However, for Evoke, where fine local search is desirable,
ESP uses only mutation.  The top quarter of the chromosomes in  each
subpopulation are duplicated and the copies are mutated by adding
Cauchy distributed noise  to all of their weight values.

The support vector method used to compute the weights ($w_{ij}$ in
equation~\ref{eq:yt})  is
a large scale approximation of the quadratic constrained optimization,
as implemented in~\cite{torch}.    

For continuous function generation,
{\em backprojection} (or {\em teacher forcing} in standard RNN
terminology) is used, 
where the predicted outputs are
fed back as inputs in the next time step:
\[
\boldsymbol{\phi}(t) =
f(\mathbf{u}(t),y(t-1),\mathbf{u}(t-1),\ldots,y(0), \mathbf{u}(0)).
\]
During training and validation,
the correct target values are  backprojected, in effect ``clamping''
 the network's outputs to the right values. During
testing, the network backprojects its own predictions.

\section{\label{sc:exp}Experimental Results}

Experiments were carried out on two test problems: context-sensitive
languages, and multiple superimposed out-of-phase sine waves.  These
tasks were chosen to highlight Evoke's ability to perform well in both
discrete and continuous domains.  The first task is of the type
standard SVMs cannot deal with at all; the second is of the type even
the  recent ESNs \cite{Jaeger:04} cannot deal with.

\subsection{Context-Sensitive Grammars}

\begin{table}
\centering
\begin{tabular}{|c|c|c|c|} \hline
Training data & G-LSTM     & PI-Evolino & Evoke   \\ \hline \hline  
1..10 & 1..29    & 1..53  & 1..257  \\ \hline  
1..20 & 1..67    & 1..95  & 1..374  \\ \hline

\end{tabular}
\label{tb:csl}
\caption{{\bf Generalization results for the $a^nb^nc^n$ language.}
Since traditional SVMs cannot solve this task at all, the table
compares Evoke to gradient-based LSTM (G-LSTM), the  only pre-2005 subsymbolic
method that has reliably learnt this problem, and pseudoinverse-based 
Evolino (PI-Evolino).  The left column shows the set of legal strings 
used to train each method.  The
other columns show the set of strings that each method was able to
accept after training.  The results for G-LSTM 
are from~\cite{Gers:01ieeetnn}, and for Evolino from
\cite{Schmidhuber:05ijcai,Wierstra:05geccoevolino}.
Average of 20 runs.}
\end{table}

Standard SVMs, or any approach based on a fixed time window, cannot
learn to recognize context-sensitive languages where the length of the
input sequence is arbitrary and unknown in advance.   For this reason
we focus on  the simplest such language, namely,
$a^nb^nc^nT$ (i.e.\ strings of $n$ $a$s, followed by $n$ $b$s,
followed by $n$ $c$s, and  ending with the termination symbol $T$).
Classifying exemplars of this language entails counting symbols and
remembering counts until the whole string has been read.  
Since traditional SVMs cannot solve this task at all, 
we compare Evoke to the pseudoinverse-based Evolino, and the only 
pre-2005 subsymbolic learning machine that has satisfactorily 
solved this problem, namely, gradient-based LSTM~\cite{Gers:01ieeetnn}.

Symbol strings were presented to the networks, one symbol at a time.
The networks had 4 input units, one for each possible symbol: $S$ for
start, $a$, $b$, and $c$.  An input is set to 1.0 when the
corresponding symbol is observed, and -1.0 when it is not present.
The network state was fed as input to four distinct SVM classifiers,
and each was trained  to predict one of the possible following symbols
$a$, $b$, $c$ and $T$.

Two sets of 20 simulations were run  each using a different training
set of legal strings, $\{a^nb^nc^n\}, n=1..N$, where $N$ was 10 and
20. The second half of each set was used for  validation, and the
fitness of each individual was evaluated as the sum of training and
validation error, to be minimized by evolution.
  
LSTM networks with 5 memory cells were evolved, with random initial values
for the weights between $-5.0$ and $5.0$.  The Cauchy
noise parameter $\alpha$ for both mutation and  burst mutation was set
to $0.1$, i.e.\ $50\%$ of the mutations is kept within this bound. In
keeping with the  setup in~\cite{Gers:01ieeetnn}, we added a bias unit
to the  Forget gates and Output gates with values of $+1.5$ and
$-1.5$, respectively.   The SVM parameters 
were chosen heuristically: a Gaussian
kernel with  standard deviation $2.0$ and capacity $100.0$.
Evolution was terminated after 50 generations,
after which the  best network in each simulation was tested.    The
results are summarized in Table~\ref{tb:csl}.

Evoke learns in approximately 6 minutes on average (on a $3$ GHz
desktop)  but, more importantly, it is able to generalize far better
than  G-LSTM---the only gradient-based RNN so far
that has achieved good generalization on such tasks
\cite{Gers:2000nc,Gers:01ieeetnn,Schmidhuber:02nc,Perez:03}.

While being superior for $N=10$ and $N=20$, the performance of Evoke
degraded for larger values of $N$, for which both PI-Evolino and G-LSTM
achieved better results.

\subsection{\label{sc:sine}Multiple Superimposed Sine Waves}

\begin{figure}[tb]
\begin{center} 
\epsfig{file=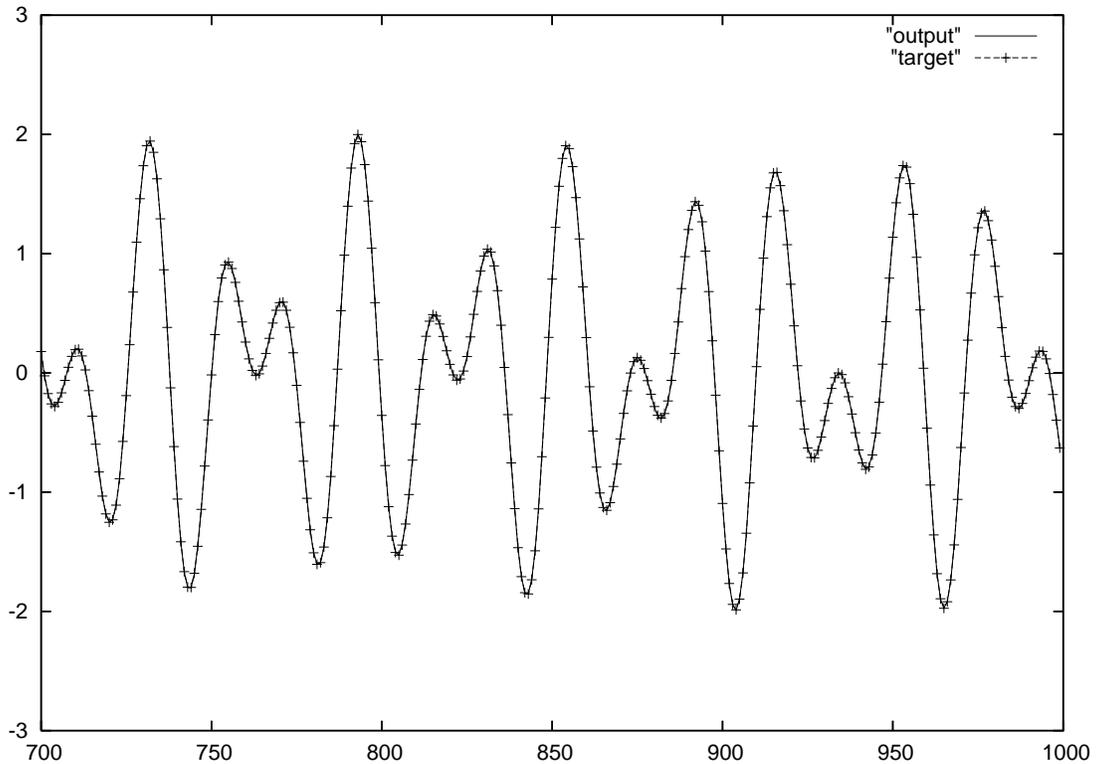, width=\linewidth}
\end{center}
\label{fig:double}
\caption{{\bf Performance of Evoke on the double superimposed sine
wave task.}  The plot shows the generated output (continuous line)
of a typical network produced after 50 generations (3000 evaluations),
compared with the test set (dashed line with crosses).
}
\end{figure}

In~\cite{jaeger:web2004}, the author reports that Echo State  Networks
\cite{Jaeger:04} are unable to learn functions composed of multiple
superimposed oscillators. Specifically, functions like
$sin(0.2x)+sin(0.311x)$, in which the individual sines have the same
amplitude but their frequencies are not multiples of each other.
G-LSTM also has difficulties  in solving such tasks
quickly.

For this task, networks with 10 memory cells
were evolved for $50$ generations to predict
$400$ time steps of the above function, 
excluding the first
$100$ as washout time;  fitness was evaluated summing the error over
the training set (points $101..400$) and a validation set (points
$401..700$), and  then tested on another set of data points from
time-steps $701..1000$. This time
the weight range was set to $[-1.0,1.0]$, and a Gaussian kernel with
standard deviation $2.0$ and capacity $10.0$ was used for the SVM.

On $20$ runs with different random seeds, the average summed squared
error over the test set ($300$ points) was $0.021$.
On the same problem, though, pseudoinverse-based Evolino reached a 
much better value of $0.003$.
Experiments with three superimposed waves, as 
in~\cite{Schmidhuber:05ijcai,Wierstra:05geccoevolino}, gave
unsatisfactory results.

Figure~\ref{fig:double} shows the behavior of one of the double sine
wave Evoke networks on the test set.

\section{\label{sc:conclusion}Conclusion}

We introduced  the first kernel-adapting, truly sequential SVM-based
classifiers and predictors.  They are trained by the Evoke algorithm:
EVOlution of systems with KErnel-based outputs. Evoke is a special
case of the recent Evolino class of algorithms
\cite{Schmidhuber:05ijcai,Wierstra:05geccoevolino} in which 
a supervised learning module (SVM in this case) is employed to assign fitness
 to  the evolving recurrent systems that pre-process inputs.
Our particular Evoke implementation uses the ESP algorithm to
coevolve the hidden nodes of an LSTM RNN.

This versatile method can
deal with long time lags between discrete events as well as with
continuous time-series prediction.  It is able to solve a
context-sensitive grammar task that standard SVMs cannot solve even in
principle. It also outperforms ESNs and previous
state-of-the-art RNN algorithms for such tasks (G-LSTM) in
terms of generalization. Finally, Evoke also quickly solves a
task involving multiple superimposed  sine waves on which ESNs
fail, and where G-LSTM is slow.

The present work represents a pilot study of evolutionary recurrent SVMs.  As for
its performance, Evoke was generally better than gradient-based LSTM,
but worse than the pseudoinverse-based Evolino
\cite{Schmidhuber:05ijcai,Wierstra:05geccoevolino}. One possible
reason for this could be that the kernel mapping of the SVM component
induces a more rugged fitness landscape that makes evolutionary
search harder. Future work will further explore Evoke's limitations, and
ways to circumvent them, including the co-evolution of SVM kernel
parameters.

\bibliographystyle{unsrt}
\bibliography{evoketr}

\begin{thebibliography}{10}

\bibitem{Schmidhuber:05ijcai}
J.~Schmidhuber, D.~Wierstra, and F.~J. Gomez.
\newblock Evolino: Hybrid neuroevolution / optimal linear search for sequence
  prediction.
\newblock In {\em Proceedings of the 19th International Joint Conference on
  Artificial Intelligence (IJCAI)}, pages 853--858. Morgan Kaufman, 2005.

\bibitem{Wierstra:05geccoevolino}
D.~Wierstra, F.~J. Gomez, and J.~Schmidhuber.
\newblock Modeling non-linear dynamical systems with {Evolino}.
\newblock In {\em Proc. GECCO 2005, Washington, D. C.}, pages 1795--1802, New
  York, 2005. ACM Press.

\bibitem{Vapnik:95}
V.~Vapnik.
\newblock {\em The Nature of Statistical Learning Theory}.
\newblock Springer, New York, 1995.

\bibitem{mukherjee97nonlinear}
S.~Mukherjee, E.~Osuna, and F.~Girosi.
\newblock Nonlinear prediction of chaotic time series using support vector
  machines.
\newblock In J.~Principe, L.~Giles, N.~Morgan, and E.~Wilson, editors, {\em
  {IEEE} Workshop on Neural Networks for Signal Processing {VII}}, page 511.
  IEEE Press, 1997.

\bibitem{uller97predicting}
K.~M\"{u}ller, A.~Smola, G.R\"{a}tsch, B.~Sch\"{o}lkopf, J.~Kohlmorgen, and
  V.~Vapnik.
\newblock Predicting time series with support vector machines, 1997.

\bibitem{Salomon:02icslp}
Jesper Salomon, Simon King, and Miles Osborne.
\newblock Framewise phone classification using support vector machines.
\newblock In {\em Proceedings International Conference on Spoken Language
  Processing}, Denver, 2002.

\bibitem{Jaakkola:98nips}
Tommi~S. Jaakkola and David Haussler.
\newblock Exploiting generative models in discriminative classifiers.
\newblock In {\em Proceedings of the 1998 conference on Advances in neural
  information processing systems II}, pages 487--493, Cambridge, MA, USA, 1999.
  MIT Press.

\bibitem{Jebara:04jmlr}
Tony Jebara, Risi Kondor, and Andrew Howard.
\newblock Probability product kernels.
\newblock {\em J. Mach. Learn. Res.}, 5:819--844, 2004.

\bibitem{Lodhi:00nips}
Huma Lodhi, John Shawe-Taylor, Nello Cristianini, and Christopher J. C.~H.
  Watkins.
\newblock Text classification using string kernels.
\newblock In {\em NIPS}, pages 563--569, 2000.

\bibitem{Shimodaira:02nips}
H.~Shimodaira, K.-I. Noma, M.~Nakai, and S.~Sagayama.
\newblock Dynamic time-alignment kernel in support vector machine.
\newblock In T.~G. Dietterich, S.~Becker, and Z.~Ghahramani, editors, {\em
  Advances in Neural Information Processing Systems 14}, Cambridge, MA, 2002.
  MIT Press.

\bibitem{Bengio:95b}
Y.~Bengio and P.~Frasconi.
\newblock Diffusion of credit in markovian models.
\newblock In G.~Tesauro, D.~S. Touretzky, and T.~K. Leen, editors, {\em
  Advances in Neural Information Processing Systems 7}, pages 553--560. MIT
  Press, 1995.

\bibitem{Hochreiter:01icann}
Sepp Hochreiter and Michael Mozer.
\newblock A discrete probabilistic memory model for discovering dependencies in
  time.
\newblock In {\em ICANN}, pages 661--668, 2001.

\bibitem{Suykens:00}
J.~A.~K. Suykens and J.~Vandewalle.
\newblock Recurrent least squares support vector machines.
\newblock {\em IEEE Transactions on Circuits and Systems-I}, 47(7):1109--1114,
  2000.

\bibitem{penrose:pseudo}
R.~Penrose.
\newblock A generalized inverse for matrices.
\newblock In {\em Proceedings of the Cambridge Philosophy Society}, volume~51,
  pages 406--413, 1955.

\bibitem{Jaeger:04}
H.~Jaeger.
\newblock Harnessing nonlinearity: Predicting chaotic systems and saving energy
  in wireless communication.
\newblock {\em Science}, 304:78--80, 2004.

\bibitem{Werbos:74}
P.~J. Werbos.
\newblock {\em Beyond Regression: New Tools for Prediction and Analysis in the
  Behavioral Sciences}.
\newblock PhD thesis, Harvard University, 1974.

\bibitem{RumelMc:86}
D.~E. Rumelhart and J.~L. McClelland, editors.
\newblock {\em Parallel Distributed Processing}, volume~1.
\newblock MIT Press, 1986.

\bibitem{Williams:89}
R.~J. Williams.
\newblock Complexity of exact gradient computation algorithms for recurrent
  neural networks.
\newblock Technical Report Technical Report NU-CCS-89-27, Boston: Northeastern
  University, College of Computer Science, 1989.

\bibitem{RobinsonFallside:87tr}
A.~J. Robinson and F.~Fallside.
\newblock The utility driven dynamic error propagation network.
\newblock Technical Report CUED/F-INFENG/TR.1, Cambridge University Engineering
  Department, 1987.

\bibitem{Pearlmutter:95}
B.~A. Pearlmutter.
\newblock Gradient calculations for dynamic recurrent neural networks: A
  survey.
\newblock {\em IEEE Transactions on Neural Networks}, 6(5):1212--1228, 1995.

\bibitem{Hochreiter:97lstm}
S.~Hochreiter and J.~Schmidhuber.
\newblock Long short-term memory.
\newblock {\em Neural Computation}, 9(8):1735--1780, 1997.

\bibitem{Rechenberg:71}
I.~Rechenberg.
\newblock {Evolutionsstrategie - Optimierung technischer Systeme nach
  Prinzipien der biologischen Evolution. Dissertation}, 1971.
\newblock Published 1973 by Fromman-Holzboog.

\bibitem{Schwefel:74}
H.~P. Schwefel.
\newblock {Numerische Optimierung von Computer-Modellen. Dissertation}, 1974.
\newblock Published 1977 by Birkh\"{a}user, Basel.

\bibitem{Holland:75}
J.~H. Holland.
\newblock {\em Adaptation in Natural and Artificial Systems}.
\newblock University of Michigan Press, Ann Arbor, 1975.

\bibitem{yao:review}
X.~Yao.
\newblock A review of evolutionary artificial neural networks.
\newblock {\em International Journal of Intelligent Systems}, 4:203--222, 1993.

\bibitem{gomez:phd}
F.~J. Gomez.
\newblock {\em Robust Nonlinear Control through Neuroevolution}.
\newblock PhD thesis, Department of Computer Sciences, University of Texas at
  Austin, 2003.

\bibitem{Gomez:03+}
F.~J. Gomez and R.~Miikkulainen.
\newblock Active guidance for a finless rocket using neuroevolution.
\newblock In {\em Proc. GECCO 2003, Chicago}, 2003.
\newblock {\em Winner of Best Paper Award in Real World Applications. Gomez is
  working at IDSIA on a CSEM grant to J. Schmidhuber.}

\bibitem{Gomez:05geccocern}
F.~J. Gomez and J.~Schmidhuber.
\newblock Co-evolving recurrent neurons learn deep memory {POMDPs}.
\newblock In {\em Proc. GECCO 2005, Washington, D. C.}, 2005.

\bibitem{Gers:2000nc}
F.~A. Gers, J.~Schmidhuber, and F.~Cummins.
\newblock Learning to forget: Continual prediction with {LSTM}.
\newblock {\em Neural Computation}, 12(10):2451--2471, 2000.

\bibitem{Gers:01ieeetnn}
F.~A. Gers and J.~Schmidhuber.
\newblock {LSTM} recurrent networks learn simple context free and context
  sensitive languages.
\newblock {\em IEEE Transactions on Neural Networks}, 12(6):1333--1340, 2001.

\bibitem{Gers:02jmlr}
F.~A. Gers, N.~Schraudolph, and J.~Schmidhuber.
\newblock Learning precise timing with {LSTM} recurrent networks.
\newblock {\em Journal of Machine Learning Research}, 3:115--143, 2002.

\bibitem{Schmidhuber:02nc}
J.~Schmidhuber, F.~Gers, and D.~Eck.
\newblock Learning nonregular languages: A comparison of simple recurrent
  networks and {LSTM}.
\newblock {\em Neural Computation}, 14(9):2039--2041, 2002.

\bibitem{Perez:03}
J.~A. P\'{e}rez-Ortiz, F.~A. Gers, D.~Eck, and J.~Schmidhuber.
\newblock Kalman filters improve {LSTM} network performance in problems
  unsolvable by traditional recurrent nets.
\newblock {\em Neural Networks}, 16(2):241--250, 2003.

\bibitem{graves:05nn}
A.~Graves and J.~Schmidhuber.
\newblock Framewise phoneme classification with bidirectional lstm and other
  neural network architectures.
\newblock {\em Neural Networks}, 18(5-6):602--610, 2005.

\bibitem{torch}
R.~Collobert, S.~Bengio, and J.~Marithoz.
\newblock Torch: a modular machine learning software library.
\newblock Technical Report 02-46, IDIAP-RR, 2002.

\bibitem{jaeger:web2004}
H.~Jaeger.
\newblock The echo state approach to recurrent neural networks, 2004.
\newblock --- seminar slides, available at
  \texttt{http://www.faculty.iu-bremen.de/hjaeger/courses/\\{S}eminar{S}pring0%
4/{ESNS}tandard{S}lides.pdf}.

\end{thebibliography}
\end{document}